\documentclass{article}

\usepackage{arxiv}

\usepackage[utf8]{inputenc} 
\usepackage[T1]{fontenc}    
\usepackage{hyperref}       
\usepackage{url}            
\usepackage{booktabs}       
\usepackage{amsfonts}       
\usepackage{nicefrac}       
\usepackage{microtype}      
\usepackage{lipsum}		
\usepackage{graphicx}
\usepackage{natbib}
\usepackage{doi}
\usepackage{amsmath}
\usepackage{amsmath}
\usepackage{amsthm}
\usepackage{booktabs}
\usepackage{algorithm}
\usepackage{algorithmic}
\usepackage[switch]{lineno}
\usepackage{enumitem}
\usepackage{subfigure}
\usepackage{diagbox}
\usepackage{amsfonts}
\usepackage{amsmath, amssymb, amsfonts}
\newcommand{\EE}{\mathbb{E}}

\newtheorem{theorem}{Theorem}
\newtheorem{assumption}{Assumption}

\newtheorem{lemma}[theorem]{Lemma}

\title{Cascading Bandits Robust to Adversarial Corruptions}

\date{} 					

\author{ {Jize Xie}\\
	Department of Industrial Engineering\\
  Hong Kong University of Science and Technology\\
	\And
	{Cheng Chen} \\
	Software Engineering Institute\\
	East China Normal University\\
	\And
    Zhiyong Wang\\
    Department of Computer Science\\
    The Chinese University of Hong Kong\\
    \And
    Shuai Li\\
    John Hopcroft Center\\
    Shanghai Jiao Tong University
}

\renewcommand{\headeright}{}
\renewcommand{\undertitle}{}
\renewcommand{\shorttitle}{}

\hypersetup{
pdftitle={A template for the arxiv style},
pdfsubject={q-bio.NC, q-bio.QM},
pdfauthor={David S.~Hippocampus, Elias D.~Striatum},
pdfkeywords={First keyword, Second keyword, More},
}

\begin{document}
\maketitle

\begin{abstract}
	Online learning to rank sequentially recommends a small list of items to users from a large candidate set and receives the users' click feedback. In many real-world scenarios, users browse the recommended list in order and click the first attractive item without checking the rest. Such behaviors are usually formulated as the cascade model. Many recent works study algorithms for cascading bandits, an online learning to rank framework in the cascade model. However, the performance of existing methods may drop significantly if part of the user feedback is adversarially corrupted (e.g., click fraud). 
   In this work, we study how to resist adversarial corruptions in cascading bandits. We first formulate the ``\textit{Cascading Bandits with Adversarial Corruptions}" (CBAC) problem, which assumes that there is an adaptive adversary that may manipulate the user feedback. Then we propose two robust algorithms for this problem, which assume the corruption level is known and agnostic, respectively. We show that both algorithms can achieve logarithmic regret when the algorithm is not under attack, and the regret increases linearly with the corruption level. The experimental results also verify the robustness of our methods.
\end{abstract}


\section{Introduction}
\label{sec:intro}
Learning to rank aims to recommend to users the most appealing ranked list of items ~\citep{kveton2015cascading,combes2015learning}. The offline learning to rank algorithms learn the ranking policy from the interaction history of users, but may face challenges in meeting the rapidly evolving user needs and preferences ~\citep{cao2007learning,trotman2005learning}. Online algorithms, on the other hand, can learn the ranking strategies from real-time user click behaviors and strive to maximize users' satisfaction during the whole learning period ~\citep{lattimore2018toprank,zhong2021thompson}. 

In many recommender systems, the user checks the recommended lists from the first item to the last, and clicks on the first item which attracts the user. Such kind of user behaviors can be modeled as the cascade model~\citep{craswell2008experimental}.
  
The cascading bandits is an online learning to rank framework in the cascade model ~\citep{kveton2015cascading}. It assumes that each item is associated with an unknown attraction probability. Under this framework, the cascading bandits algorithms recommend a list of $K$ items out of $L$ candidate items to the users, and subsequently observe the index of items clicked by the user. If the user clicks on any item in the list, the agent will receive a reward of one. Otherwise, the agent will receive a reward of zero. The goal of cascading bandits algorithms is to minimize the cumulative regret, which represents the difference between the algorithms' cumulative reward and the cumulative reward of always recommending the optimal list over the whole $T$ time steps.

The cascading bandits have been widely studied recently~\citep{kveton2015cascading,zhong2021thompson,kveton2015combinatorial,combes2015learning,li2019cascading}. Classical methods can achieve $O(\log(T))$ regret with stochastic user feedback. However, in some scenarios, the user feedback may be corrupted by adversarial attacks, which may affect the performance  of the learning algorithms~\citep{lykouris2018stochastic,gupta2019better}. 
A typical example is the click fraud in online advertising, where a bot can fake user clicks on some ads to deceive the learning algorithm. The bot may repeatedly make searches that trigger a certain ad and ignore it, making it appear that the ad has a very low click-through rate and giving its rival an advantage~\citep{lykouris2018stochastic}. Such corruptions may seriously harm the user satisfaction and the platform revenue.

Most existing cascading bandits algorithms for stochastic environments can be vulnerable to adversarial corruptions because their performance highly depends on the accurate estimation of the attraction probabilities of items, which can be destroyed by the adversarial corruptions. On the other hand, the adversarial cascading bandits algorithms, such as Ranked Bandits Algorithm (RBA)~\citep{radlinski2008learning}, can resist adversarial corruptions, but can only achieve an $O(\sqrt{T})$ regret bound in the stochastic environments. Therefore, how to design stochastic cascading bandits algorithms which are robust to the adversarial corruptions as well as achieve $O(\log(T))$ regret is an important problem. 

While lots of efforts have been made to study robust bandit algorithms in various settings including multi-armed bandits (MAB), linear bandits and combinatorial semi-bandits~\citep{lykouris2018stochastic,liu2021cooperative,lu2021stochastic,ding2022robust,wang2024online}, none of these works are proposed for the cascading bandits. The main challenge in designing a robust cascading bandits algorithm stems from the requirement of identifying the
$K$ instead of one optimal item under possibly partial feedback as the user may not check the whole recommended list. 

In this paper, we study the “\textit{Cascading bandits with adversarial corruptions}” (CBAC) problem where exists an adversary who can manipulate the user feedback. We propose two robust algorithms called \emph{CascadeRKC} and \emph{CascadeRAC} for this problem. Both algorithms are designed upon the novel position-based elimination (PBE) algorithm for the cascade model, which is an extension of active arm elimination (AAE)~\citep{even2006action,lykouris2018stochastic} for multi-armed bandits. The \emph{CascadeRKC} algorithm requires to
know the corruption level but has better performance, while the \emph{CascadeRAC} algorithm can resist agnostic corruptions with a slightly larger regret upper bound. We show that both algorithms can achieve gap-dependent logarithmic regret when the algorithm is not under attack, and the regret increases linearly with the corruption level. We also conduct extensive experiments in various datasets and settings. The empirical results show that our algorithms are robust to different kinds and levels of corruptions.

\section{Relatet Works}
Our work is closely related to two research lines: cascading bandits and bandits robust to adversarial corruptions.
\subsection{Cascading Bandits}
\citet{kveton2015cascading} and \citet{combes2015learning} first formulate the cascading bandits, which consider online learning to rank under the cascade model. 
\citet{kveton2015cascading} propose two algorithms called CascadeUCB1 and CascadeKL-UCB to solve the cascading bandits problem, while \citet{combes2015learning} consider a more general model and propose PIE and PIE-C algorithms for it. \citet{kveton2015combinatorial} define the combinatorial cascading bandits where the learning agent only gets a reward if the weights of all chosen items are one, and it proposes the CombCascade algorithm to solve this problem with both gap-dependent and gap-free regret upper bounds. \citet{zoghi2017online} and \citet{lattimore2018toprank} study the generalized click model and provide gap-dependent regret bounds and gap-free regret bounds, respectively. Recently, \citet{vial2022minimax} provide matching upper and lower bounds for the gap-free regret for the case of unstructured rewards. \citet{zhong2021thompson} establish regret bounds for Thompson sampling cascading bandit algorithms, which are slightly worse than UCB-based methods. Another line of works studies the contextual cascading bandits where the feedback depends on the contextual information~\citep{li2016contextual, li2018online, zong2016cascading,li2019online, li2020cascading}. Some other variants of cascading bandits are also studied, such as the cascading non-stationary bandits \citep{li2019cascading},   which are beyond this work's scope. However, none of these works consider the potential adversarial corruptions. 

\subsection{Bandits Robust to Adversarial Corruptions}
\citet{lykouris2018stochastic} first study the multi-armed bandits problem with adversarial corruptions bounded with the corruption level $C$ and propose two robust elimination-based algorithms. Then \citet{gupta2019better} further improve the algorithms in \citep{lykouris2018stochastic} and give a better regret bound. \citet{kapoor2019corruption} develop algorithms robust to sparse corruptions in both the multi-armed bandits setting and the contextual setting, but their algorithms are not general enough for certain corruption mechanisms. \citet{zimmert2021tsallis} use an online mirror descent method with Tsallis entropy. \citet{lu2021stochastic} study the stochastic graphical bandits with adversarial corruptions, where an agent has to learn the optimal arm to pull from a set of arms that are connected by a graph. \citet{agarwal2021stochastic} study the stochastic dueling bandits with adversarial corruptions where the arms are compared pairwise. And \citet{liu2021cooperative} first consider corruptions in a multi-agent setting and manage to minimize the regret as well as maintain the communication efficiency. \citet{hajiesmaili2020adversarial} consider the corruptions for the adversarial bandits. There are also some works studying the stochastic linear bandits with adversarial corruptions, including \citep{li2019stochastic,ding2022robust,he2022nearly,dai2024online}. None of these works consider the cascading bandits setting.

To the best of our knowledge, our work is the first one to study cascading bandit algorithms robust to adversarial corruptions, with both rigorous theoretical guarantees and convincing experimental results.

\section{Problem Setup}
\label{sec:set}
In this section, we present the problem formulation for the CBAC problem. We begin by defining the item set $\boldsymbol{E} = \{1, 2, \dots, L \}$ of $L$ ground items. Each item $a \in \boldsymbol{E}$ is associated with an unknown attraction probability $w(a) \in [0, 1]$. Let $K < L$ be the number of positions in the list. Without loss of generality, we assume that $w(1)\ge w(2) \ge \dots \ge w(K) > w(K+1) \ge \dots \ge w(L)$, and each item $a_{k}$ in the list attracts the user with the probability $w(a_{k})$ independently. We use $\Delta_{i,j}=w(a_i)-w(a_j)$ to represent the attraction probability difference between two items $i$ and $j$, and $\Pi_{K}(\boldsymbol{E})$ to denote the set of all K-permutations of the ground set $\boldsymbol{E}$. Let $\boldsymbol{R}_{t} \in \{0, 1\}^{L} $ denote the attraction indicator of the $t$-th round, i.e., $\boldsymbol{R}_{t}(i)=1$ means that the item $i$ is attractive at round $t$. And $\boldsymbol{R}_{t}$ is drawn i.i.d from $P$, a probability distribution over a binary hypercube $\{0,1\}^L$. The clicked item can be represented by $\boldsymbol{Y}_{t} = \inf\{k \in [K]: \boldsymbol{R}_{t}(a_{k}) = 1\}$. If no item is clicked, then $\boldsymbol{Y}_{t} = \infty$. 

The protocol of the CBAC problem  at each round $t\in[T]$ is described as follows:
\begin{itemize}[leftmargin=0.5cm]
    \item The agent recommends a list of $K$ items $\boldsymbol{A}_{t} = (a_{1}, a_{2}, \dots, a_{K}) \in \Pi_{K}(\boldsymbol{E})$ to the user.
    \item The user examines the list from the first item to the last. If item $a_{k}$ is attractive, the user clicks on $a_{k}$ and does not examine the rest items. The reward is one if and only if the user is attracted by at least one item in $\boldsymbol{A}_{t}$, i.e., the reward function can be written as:
    \begin{equation}
    \label{reward}
    f(\boldsymbol{A}_{t}, \boldsymbol{R}_{t}) = 1 - \prod_{a_k \in \boldsymbol{A}_{t}}(1 - \boldsymbol{R}_{t}(a_{k})). 
    \end{equation}
    \item The attacker observes the recommended list and  $\boldsymbol{R}_{t}$, and designs a corrupted feedback $\boldsymbol{\Tilde{{R}}_{t}}$ and $\boldsymbol{\Tilde{Y}}_{t
    }$. 
    \item The agent receives $\boldsymbol{\Tilde{Y}}_{t
    }$.
\end{itemize}
We assume that the adversary can be ``adaptive", i.e., it can decide whether to corrupt the
user feedback according to the previous lists and clicks. We say that the problem instance is $C$-corrupted if the total corruption is at most $C$:
\begin{equation*}
    \sum_{t=1}^{T}\max_{a_k\in \boldsymbol{A_t}}\lvert \boldsymbol{R}_{t}(a_k) - \Tilde{\boldsymbol{{R}}}_{t}(a_k) \rvert \le C.
\end{equation*}

Similar to \cite{kveton2015cascading}, we make the following assumption:
\begin{assumption}
\label{assump1}
The attraction indicators in the ground set are distributed as:
\begin{equation*}
    P(\boldsymbol{R}) = \prod_{a \in \boldsymbol{E}}P_{a}(\boldsymbol{R}(a)).
\end{equation*}
where $P_{a}$ is a Bernoulli distribution with mean $w(a)$.
\end{assumption}
With this assumption and the fact that in the cascade model the attraction probability $w(a)$ only depends on $a$ and is independent of other items, the expected reward of any action $A$ can be written as:
$\mathbb{E}[f(\boldsymbol{A}, \boldsymbol{R})]=f(\boldsymbol{A}, \boldsymbol{w}_A)$, where $\boldsymbol{w}_A$ represents the vector consisted of the items' attraction probabilities in $\boldsymbol{A}$. Then cumulative regret is defined as:
\begin{equation}
    \label{nips:regret}
    R(T) = \sum_{t=1}^{T}(f(\boldsymbol{A}^*, \boldsymbol{R}_t) - f(\boldsymbol{A}_{t}, \boldsymbol{R}_{t})),
\end{equation}
where $\boldsymbol{A}^{*}$ represents the optimal permutation. The goal of the agent is to minimize the cumulative regret.
\section{Algorithms}
\label{sec:alg}
\begin{algorithm}[t]
    \caption{\emph{Position-based Elimination}}
    \label{alg:pbe}
\begin{minipage}{\columnwidth}
\begin{algorithmic}[1]
    \STATE{\textbf{Initialization}: $T(a)=0$, $\hat{w}(a)=0$, $\boldsymbol{M}_{k} = \emptyset$ for all  $a \in \boldsymbol{E}$, $k \in [1, \ldots, K]$, and $\boldsymbol{A}_{t} = \emptyset$. }
    \FORALL{$t=1,2,\ldots,T$}
    \STATE{Initialize $\boldsymbol{A}_{t} = \emptyset$}
    \FOR{position $k = 1, 2, \ldots, K$}
    \STATE{Select an item $a_{k}$ from $\boldsymbol{E} \setminus (\boldsymbol{M}_{k} \cup \boldsymbol{A}_{t})$ with smallest $T(a)$}.
    \STATE{Add item $a_{k}$ to $\boldsymbol{A}_{t}$.}
    \ENDFOR
    \STATE{Display $\boldsymbol{A}_{t}$ to the user and observe click $\boldsymbol{Y}_{t}$.}
    \FOR{$k = 1, 2, \ldots, \min\{\boldsymbol{Y}_{t}, K\}$}
\STATE Compute $\hat{w}(a_{k}) = \frac{T(a_{k}) \times  \hat{w}(a_{k}) +  \mathbb{I}\{\boldsymbol{Y}_{t} = k\}}{T(a_{k}) + 1}$.
\STATE $T(a_{k}) = T(a_{k}) + 1$.
\ENDFOR
\FOR{$a_{k} \in \boldsymbol{A}_{t}$}
\IF{exist $k$ items satisfy $\hat{w}(a)-\hat{w}(a_k) \ge wd(a) + wd(a_{k})$}
\STATE{Eliminate $a_k$ and add it to $\boldsymbol{M}_{k}$.}
\ENDIF
\ENDFOR
    \ENDFOR
\end{algorithmic}   
\end{minipage}
\end{algorithm}

\begin{algorithm}[htb]
    \caption{Cascading bandits robust to known corruptions (\emph{CascadeRKC}) }
    \label{alg:known}
\begin{minipage}{\columnwidth}
\begin{algorithmic}[1]
\STATE{\textbf{Initialization}: $T^{\ell}(a) = 0$, $\hat{w}^{\ell}(a) = 0$, $\boldsymbol{M}^{\ell}_{k} = \emptyset$ for all $a \in \boldsymbol{E}$, $k \in [1, \ldots, K]$, $\ell \in \{F, S\}$, and $\boldsymbol{A}_{t} = \emptyset$}
\FORALL{$t=1,2,\ldots, T$}
\STATE{Initialize $\boldsymbol{A}_{t} = \emptyset$}
\STATE{Run instance $\ell =S$ with probability $1/C$, else run instance $\ell =F$.}
\FOR{position $k = 1, 2, \ldots, K$}
\IF{$\boldsymbol{E} \setminus (\boldsymbol{M}_{k}^{\ell} \cup \boldsymbol{A}_{t}) \neq \emptyset$}
\STATE{Select an item $a_{k}$ from $\boldsymbol{E} \setminus (\boldsymbol{M}_{k}^{\ell} \cup \boldsymbol{A}_{t})$ with smallest $T^{\ell}(a)$}.
\ELSE
\STATE{Select an arbitrary item $a_{k}$ from $\boldsymbol{E} \setminus (\boldsymbol{M}_{k}^{S} \cup \boldsymbol{A}_{t})$}.
\ENDIF
\STATE{Add  item $a_{k}$ to $\boldsymbol{A}_{t}$}.
\ENDFOR
\STATE{Display $\boldsymbol{A}_{t}$ to the user and observe click $\boldsymbol{\Tilde{Y}}_{t}$}.
\FOR{$k = 1, 2, \ldots, \min\{\boldsymbol{\Tilde{Y}}_{t}, K\}$}
\IF{$a_{k} \in \boldsymbol{E} \setminus \boldsymbol{M}_{k}^{\ell}$}
\STATE Compute $\hat{w}^{\ell}(a_{k}) = \frac{T^{\ell}(a_k) \times  \hat{w}^{\ell}(a_{k}) +  \mathbb{I}\{\boldsymbol{\Tilde{Y}}_{t} = k\}}{T^{\ell}(a_{k}) + 1}$.
\STATE $T^{\ell}(a_{k}) = T^{\ell}(a_{k}) + 1$.
\ENDIF
\ENDFOR
\FOR{$k \in [1, \dots, K]$}
\FOR{item $a_{i} \in \boldsymbol{E} \setminus \boldsymbol{M}_{k}^{\ell}$}
\IF{exist $k$ items satisfy $ \hat{w}^{\ell}(a) -  \hat{w}^{\ell}(a_{i}) \ge wd^{\ell}(a)  + wd^{\ell}(a_i)$ }
\STATE{Eliminate $a_{i}$ and add it to $\boldsymbol{M}_{k}^{\ell}$}.
\IF{$\ell = S$}
\STATE{Also add $a_{i}$ to $\boldsymbol{M}_{k}^{F}$}.
\ENDIF
\ENDIF
\ENDFOR
\ENDFOR
\ENDFOR
\end{algorithmic}
\end{minipage}
\end{algorithm}

In this section, we study robust algorithms for the CBAC problem. Previous robust methods for the multi-armed bandits (MAB) problem \cite{lykouris2018stochastic,gupta2019better} resist the corruptions by maintaining multiple active arm elimination (AAE) 
instances, which removes the sub-optimal items based on specific criteria. The AAE method maintains a single elimination set with uniform elimination rules for all items and can only recommend one item as it is designed for MAB. In cascading bandits, a direct implementation of AAE is to relax the elimination rules to recommend $K$ items. However, the effectiveness of AAE is closely tied to its elimination rules; if these rules are too lenient, the process of eliminating sub-optimal items is delayed. Consequently, sub-optimal items may appear frequently before elimination, leading to a large cost. 

To address this challenge, we first design a novel position-based elimination method (PBE) tailored for cascading bandits in Section \ref{sec:pbm}. The key idea of the PBE algorithm is to maintain one elimination set for each position with strict elimination rules. This approach effectively controls the occurrences of sub-optimal items. We then combine the proposed PBE algorithm with the idea of maintaining multiple instances from \cite{lykouris2018stochastic} to tackle the CBAC problem with known and agnostic corruption levels, as introduced in Section \ref{ssec:rkc} and \ref{sec:rac}.

\subsection{The Position-based Elimination Algorithm} \label{sec:pbm}

In the PBE algorithm, we maintain one set $\boldsymbol{M}_k$ for each position $k$ to track the eliminated items. An item $a_{i}$ will be eliminated when there exist another $k$ items whose lower confidence bound (LCB, empirical estimation minus confidence radius) is larger than the upper confidence bound (UCB, empirical estimation plus confidence radius) of $a_{i}$. The confidence radius $wd(a)$ of item $a$ is in the order of $O\left(\sqrt{\frac{\log(T)}{T(a)}}\right)$. The eliminated items will be added to $\boldsymbol{M}_{k}$ and these items will not be recommended by the agent at this position $k$ in the future. In this way, we expect that eventually there will be $k$ available items for the position $k$, and since the same item cannot appear in different positions at one round, the agent can finally make accurate recommendations. The PBE method can effectively limit the times that sub-optimal items replace high attraction probability optimal items and consequently achieve a logarithmic regret bound for the cascading bandits. We present the details of the PBE algorithm in Algorithm \ref{alg:pbe}.

\subsection{The \emph{CascadeRKC} Algorithm}
\label{ssec:rkc}

To design robust cascading bandits algorithms based on PBE, we first need to ensure that the adversarial corruptions cannot easily make the optimal items obsolete, which can be achieved by enlarging the confidence radius of PBE. Moreover, to keep our algorithms effective in stochastic environments, we need to limit the delayed elimination of sub-optimal items caused by the enlarged confidence radius, i.e., to avoid the sub-optimal items being played too many times. Thus, we propose to maintain two instances of PBE algorithms simultaneously and equip the faster instance $F$ with the normal confidence radius, the slower instance $S$ with the enlarged confidence radius. And we
sample the slower instance with probability $1/C$ at each round $t$ such that the sub-optimal items in $S$ can't appear too many times when the input is stochastic.
 Moreover, when the corruption level is $C$, the expected amount of corruption that falls in the slower instance $S$ will be a constant. Thus, the feedback of the slower instance will be nearly stochastic and the influence of corruption can be much milder. In terms of the selection of the enlarged confidence radius, as the actual corruption in the $S$ instance will be no more than $O\left(\log\left(T\right)\right)$ (see Lemma \ref{lemma:slow corruption}), so we choose $wd^{S}(a)=O\left(\sqrt{\frac{\log(T)}{T^{S}(a)}} +\frac{\log\left(T\right)}{T^{S}(a)} \right)$. Furthermore, when the slower instance $S$ eliminates an item, the faster instance $F$ also eliminates it. This connection ensures that the faster instance remains efficient while inheriting the ability to eliminate sub-optimal items of the slower instance.

We present the details of the \emph{CascadeRKC} algorithm in Algorithm \ref{alg:known}. At each round $t$, the agent first samples an instance $\ell$ between the $F$ and $S$ instances (Line 4).
Then the agent decides the recommended list $\boldsymbol{A}_{t}$ position by position. If the available item set $\boldsymbol{E} \setminus (\boldsymbol{M}_{k}^{\ell} \cup \boldsymbol{A}_{t})$ for position $k$ is not empty, then select one item $a_{k}$ with the smallest played times. Here $\boldsymbol{M}_{k}^{\ell}$ is used to keep track of the eliminated items for position $k$ in the selected instance $\ell$. Notice that the items in $F$ may be all eliminated, then the agent needs to play an arbitrary item from the slower instance (Line 6-10). The selected item will be added to $\boldsymbol{A}_{t}$ so that the agent cannot select this item for the following positions. Once the recommended list $\boldsymbol{A}_t$ is determined, the agent observes the user feedback and updates the statistics accordingly. 
Finally, the agent checks if there is any item that should be eliminated and adds the eliminated items to $\boldsymbol{M}_{k}^{\ell}$ (Line 19-28) according to the PBE rules.

\subsection{The \emph{CascadeRAC} Algorithm}
\label{sec:rac}
\begin{algorithm}[tb]
    \caption{Cascading bandits robust to agnostic corruptions (\emph{CascadeRAC})}
    \label{alg:agnostic}
\begin{minipage}{\columnwidth}
\begin{algorithmic}[1]
\STATE{\textbf{Initialization}: $T^{\ell}(a) = 0$, $\hat{w}^{\ell}(a) = 0$, $\boldsymbol{M}^{\ell}_{k} = \emptyset$ for all $a \in \boldsymbol{E}$, $k \in [1, \ldots, K]$, $\ell \in [\log(T)]$, and $\boldsymbol{A}_{t} = \emptyset$}
\FORALL{$t=1,2,\ldots, T$}
\STATE{Initialize $\boldsymbol{A}_{t} = \emptyset$}.
\STATE{Run instance $\ell \in \log(T)$ with probability $2^{-\ell}$, with remaining probability, $\ell=1$.}
\FOR{position $k = 1, 2, \ldots, K$}
\IF{$\boldsymbol{E} \setminus (\boldsymbol{M}_{k}^{\ell} \cup \boldsymbol{A}_{t}) \neq \emptyset$}
\STATE{Select an item $a_{k}$ from $\boldsymbol{E} \setminus (\boldsymbol{M}_{k}^{\ell} \cup \boldsymbol{A}_{t})$ with smallest $T^{\ell}(a)$}
\ELSE
\STATE{Select an arbitrary item $a_{k}$ from $\boldsymbol{E} \setminus (\boldsymbol{M}_{k}^{\ell^{'}} \cup \boldsymbol{A}_{t})$}, where $\ell^{'}$ is the minimum instance satisfying $\boldsymbol{E} \setminus (\boldsymbol{M}_{k}^{\ell^{'}} \cup \boldsymbol{A}_{t}) \neq \emptyset$.
\ENDIF
\STATE{Add the item $a_{k}$ to $\boldsymbol{A}_{t}$}.
\ENDFOR
\STATE{Display $\boldsymbol{A}_{t}$ to the user and observe click $\boldsymbol{\Tilde{Y}}_{t}$}
\FOR{$k = 1, 2, \ldots, \min\{\boldsymbol{\Tilde{Y}}_{t}, K\}$}
\IF{$a_{k} \in \boldsymbol{E} \setminus \boldsymbol{M}_{k}^{\ell}$}
\STATE{Compute}
    $\hat{w}^{\ell}(a_{k}) = \frac{T^{\ell}(a_k) \times  \hat{w}^{\ell}(a_{k}) +  \mathbb{I}\{\boldsymbol{\Tilde{Y}}_{t} = k\}}{T^{\ell}(a_{k}) + 1}$.
\STATE $T^{\ell}(a_{k}) = T^{\ell}(a_{k}) + 1$.
\ENDIF
\ENDFOR
\FOR{$k \in [1, \dots, K]$}
\FOR{item $a_{i} \in \boldsymbol{E} \setminus \boldsymbol{M}_{k}^{\ell}$}
\IF{exist $k$ items satisfy $ \hat{w}^{\ell}(a) -  \hat{w}^{\ell}(a_{i}) \ge wd^{\ell}(a)  + wd^{\ell}(a_i)$ }
\STATE{Eliminate $a_{i}$ and add it to $\boldsymbol{M}_{k}^{\ell}$}.
\STATE{Eliminate $a_{i}$ for all $\ell^{'} \le \ell$}.
\ENDIF
\ENDFOR
\ENDFOR

\ENDFOR
\end{algorithmic}
\end{minipage}
\end{algorithm}
The design of the \emph{CascadeRAC} algorithm is smoothly extended from the \emph{CascadeRKC} algorithm with some improvement to resist agnostic corruptions. We present the \emph{CascadeRAC} algorithm in Algorithm \ref{alg:agnostic}. Unlike \emph{CascadeRKC} where we only keep two instances to defend the known adversarial corruptions level, we have to keep $\log(T)$ instances in the \emph{CascadeRAC} algorithm since it aims to resist agnostic corruptions. Each instance $\ell$ is slower and more robust than the previous one with the enlarged confidence radius $wd^{\ell} = O\left(\sqrt{\frac{\log(T)}{T^{\ell}(a)}} + \frac{\log(T)}{T^{\ell}(a)}\right)$. And at each round, each instance $\ell$ is played with the probability $2^{-\ell}$. Similar to \emph{CascadeRKC}, if the total corruption level is $C$, then the layers $\ell > \log(C)$ will suffer a constant corruption in expectation. And if one item is eliminated in instance $\ell$, then all instances satisfy $\ell^{'} \le \ell$ also need to eliminate this item. 
Moreover, if the available item set of the sampled instance is empty, the agent needs to check the following instances until it can select an item. 
\section{Theoretical Analysis}
\label{sec:analysis}
In this section, we give the theoretical results of our algorithms. The detailed proofs are deferred to the Appendix.
\subsection{Regret Analysis of CascadeRKC}
We first give the lemma which captures the highest corruption observed by the instance $S$ in \textit{CascadeRKC}:
\begin{lemma}
\label{lemma:slow corruption}
    With probability at least $1-\delta$, when sampled with probability $1/C$, the corruption $C_{S}$ of $S$ instance in CascadeRKC can be bounded by $\log(1/\delta) + 3$ during its exploration phase.
\end{lemma}

With Lemma~\ref{lemma:slow corruption}, we can give the following lemma which bounds the rounds that a sub-optimal item $e$ is placed at the position $k$ in the  instance $S$:
\begin{lemma}
\label{lemma:slow times}
For CascadeRKC with confidence intervals $wd^{F}(a) = \sqrt{\frac{\log\left(8LT/\delta\right)}{T^{F}(a)}}$ and $wd^{S}(a)=\sqrt{\frac{\log\left(8LT/\delta\right)}{T^{S}(a)}} + \frac{2\log\left(8LT/\delta\right)}{T^{S}(a)}$, with probability at least $1-\delta_{2}$, the optimal items will never be eliminated, and a sub-optimal item $e$ will be eliminated for the position $k$ when:
\begin{equation}
\label{eq:slow times}
    T^{S}(e) \le \frac{18\log\left(8LT/\delta\right)}{\Delta_{e, k}^{2}} .
\end{equation}
\end{lemma}

With Lemma~\ref{lemma:slow times}, we can get the corresponding rounds that the sub-optimal item $e$ is placed at position $k$ in the instance $F$. Further we can give the following theorem which bounds the cumulative regret of \textit{CascadeRKC}. 
\begin{theorem}
For CascadeRKC with confidence intervals $wd^{F}(a) = \sqrt{\frac{\log\left(8LT/\delta\right)}{T^{F}(a)}}$ and $wd^{S}(a)=\sqrt{\frac{\log\left(8LT/\delta\right)}{T^{S}(a)}} + \frac{2\log\left(8LT/\delta\right)}{T^{S}(a)}$, with probability at least $1-\delta$, the regret upper bound for $T$ rounds satisfies:
\begin{equation}
\label{thm:known}
    R(T) \le O\left(\sum_{e=K+1}^{L}\frac{1}{\Delta_{e,K}} KLC\left(\log\left(LT/\delta\right)\right)^2\right).\notag
\end{equation}
\end{theorem}
\subsection{Regret Analysis of CascadeRAC}
The analysis of CascadeRAC is similar to CascadeRKC, we give the following theorem to bound the cumulative regret of \emph{CascadeRAC}:

\begin{theorem}
    For the CascadeRAC algorithm with $wd^{l}(a) = \sqrt{\frac{\log\left(4LT\log T/\delta\right)}{T^{l}(a)}}  + \frac{\log\left(4LT\log T/\delta\right)}{T^{l}(a)}$, the regret upper bound for $T$ rounds satisfies:
{\small
\begin{equation}
    \label{thm:agnostic}
    R(T) \le O\left(\sum_{e=K+1}^{L}\frac{K\left(LC\log\left(LT/{\delta}\right) {+} \log(T)\right)\log\left({LT}/{\delta}\right)}{\Delta_{e, K}}\right).\notag
\end{equation}
}
with probability at least $1-\delta$.
\end{theorem}

\subsection{Discussions}
We compare our theoretical results with the degenerated robust multi-armed bandit algorithms~\cite{lykouris2018stochastic} to show the tightness of  our results. 
\begin{itemize}
    \item \textbf{Known Corruption Level Case:} When $K=1$ our setting will degenerate to MAB with binary feedback and the regret bound of the CascadeRKC algorithm will be $O\left(\sum_{e \neq e^{*}}\frac{1}{\Delta_{e, e^{*}}}LC\left(\log\left(LT/\delta\right)\right)^{2}\right)$, which is exactly the same as the regret bound of the known corruption level case in \cite{lykouris2018stochastic}. 
    \item \textbf{Agnostic Corruption Level Case:} Similarly, when $K=1$, the regret bound of the CascadeRAC algorithm will be $ O\Big(\sum_{e \neq e^{*}}\frac
{1}{\Delta_{e, e^{*}}}(LC\log(LT/\delta) + \log(T))(\log({LT}/{\delta}))\Big)$, which matches the result of the agnostic corruption level case in the work \cite{lykouris2018stochastic}.
\end{itemize}
In addition, the lower bound of this problem is still unknown. However, MAB with corruptions has a lower bound $\Omega(C)$ (Theorem 4 of \cite{lykouris2018stochastic}). Thus our result is tight with regard to $C$.
\section{Experiments}
\label{Experiments}
\begin{figure}[t]
    \centering
    \includegraphics[scale=0.35]{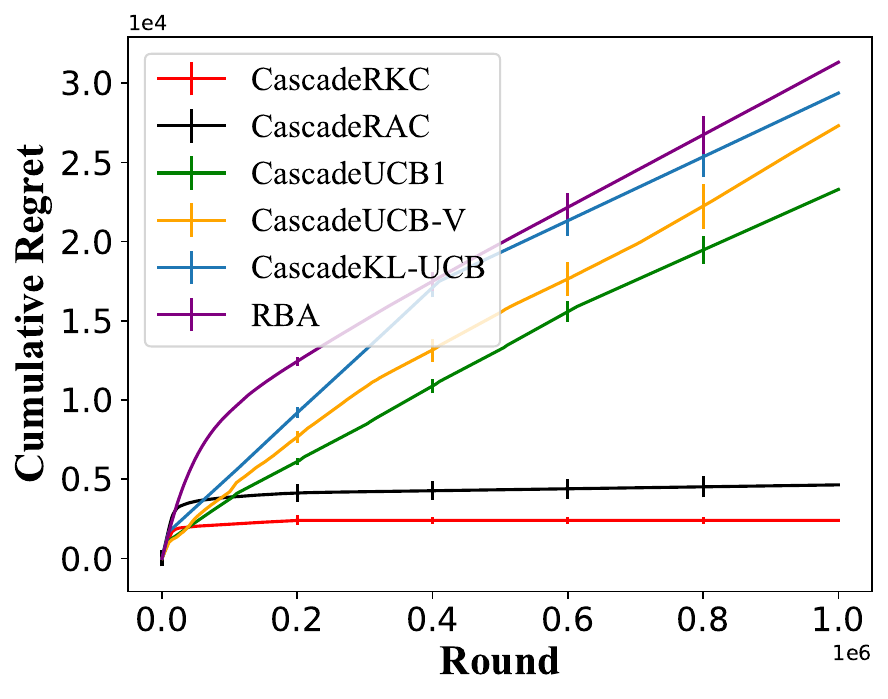}
    \caption{Comparison of cumulative regrets on the synthetic dataset with $L=500$ and $K=5$.}
    \label{fig:synthetic}
\end{figure}
\begin{figure*}[t]
\centering
    \subfigure[Yelp ]{
    \includegraphics[scale=0.33]{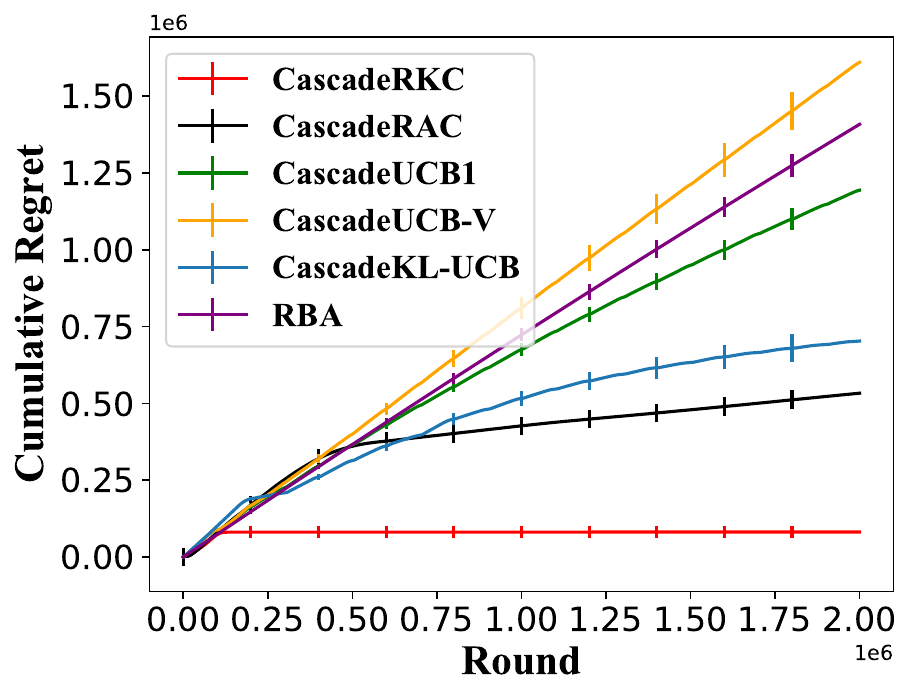}
    }
    \hfill
    \subfigure[Movielens]{
    \includegraphics[scale=0.33]{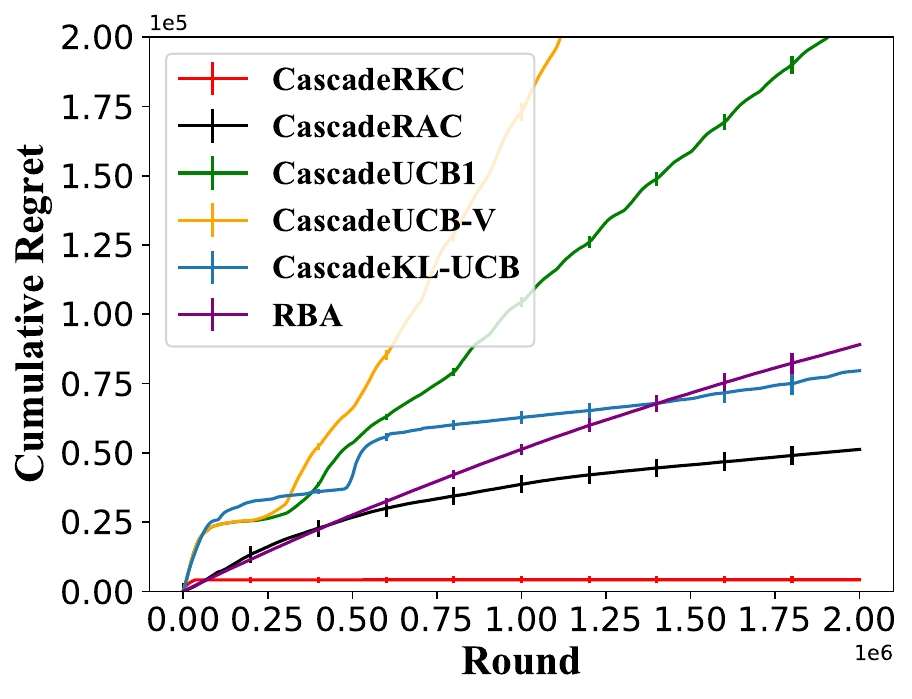}
    }
    \hfill
    \subfigure[Yandex]{
    \includegraphics[scale=0.33]{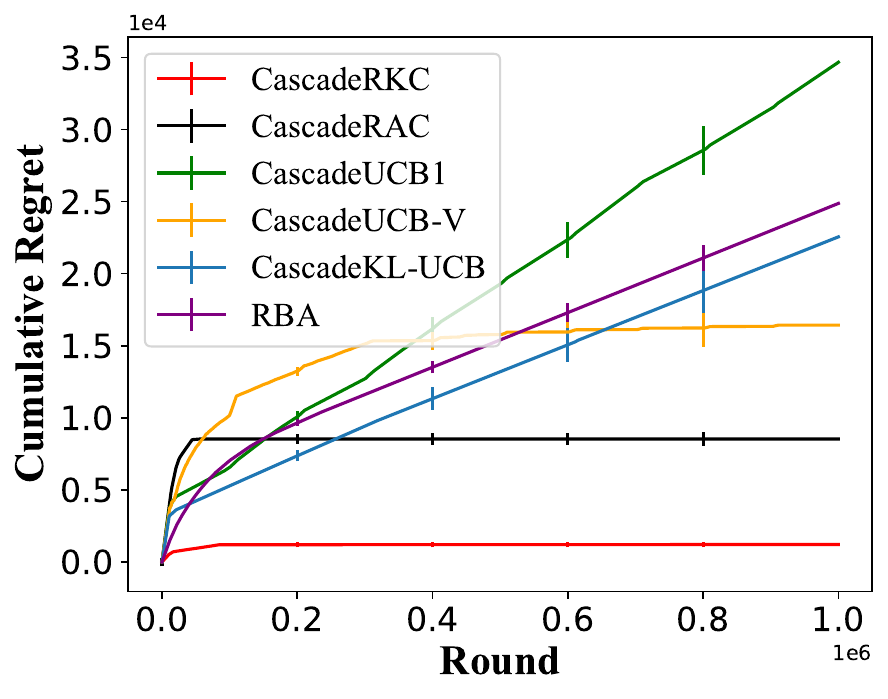}
    }
    \caption{Cumulative regret on the real datasets. (a) shows the results in the Yelp dataset, (b) shows the results in the Movielens dataset, and (c) shows the results in the Yandex dataset.}
    \label{fig: real regret}
\end{figure*}
\begin{figure*}[ht]
\centering
     \subfigure[$\Delta = 0.1$]{
    \includegraphics[scale=0.333]{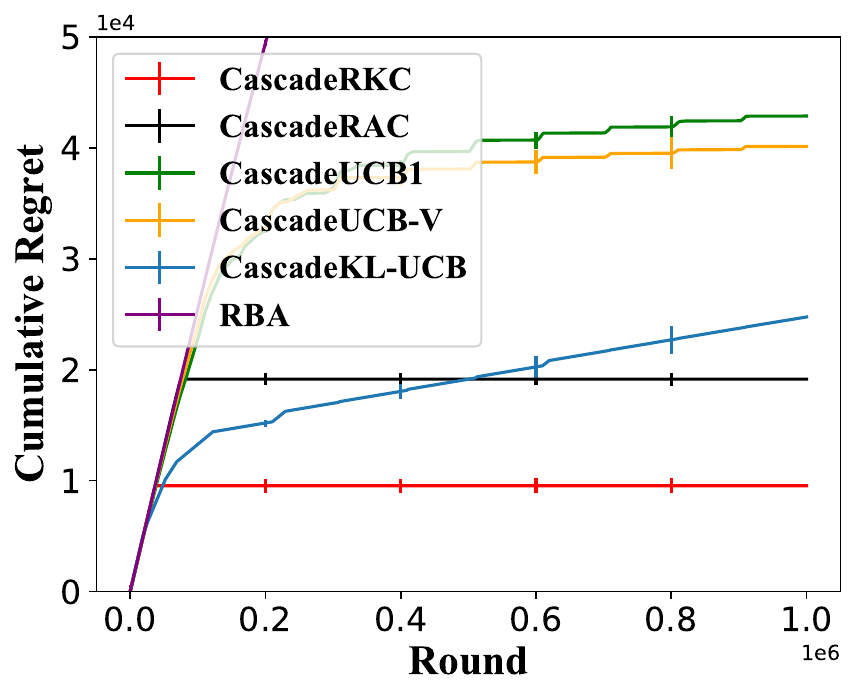}
    }
    \hfill
     \subfigure[$\Delta = 0.2$]{
    \includegraphics[scale=0.333]{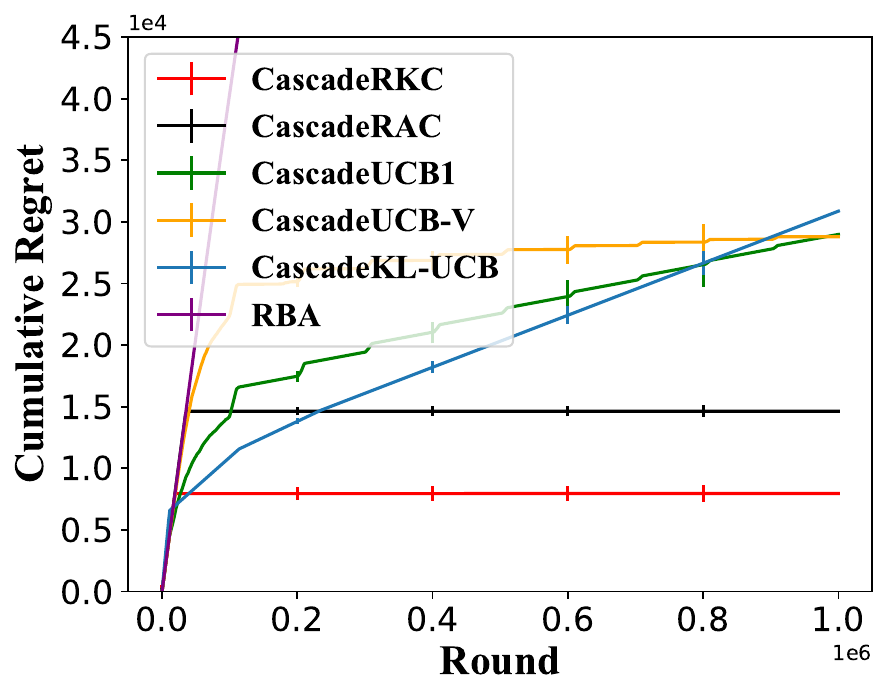}
    }
    \hfill
    \subfigure[$\Delta = 0.4$]{
    \includegraphics[scale=0.333]{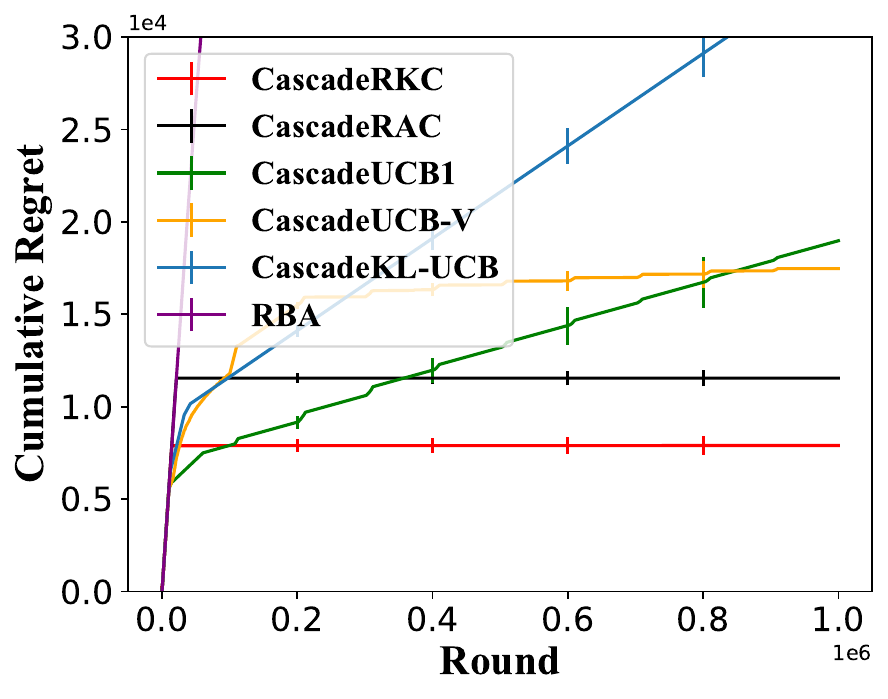}
    }
    \caption{Cumulative regret under different $\Delta$ between optimal items and the target item. (a) exhibits the results when $\Delta=0.1$, (b) exhibits the results when $\Delta=0.2$, and (c) exhibits the results when $\Delta=0.4$.}
    \label{fig: synthetic delta}
\end{figure*}

In this section, we conduct experiments in both synthetic dataset and real-world datasets to examine the performance of our algorithms. We compare our proposed CascadeRKC and CascadeRAC with CascadeUCB1~\cite{kveton2015cascading}, CascadeUCBV~\cite{vial2022minimax}, CascadeKL-UCB~\cite{kveton2015cascading}, and the Randked Bandits Algorithm (RBA) ~\cite{radlinski2008learning}. Here, the first three baselines are all UCB-based cascading bandits algorithms, and they only differ in the choices of the computation of UCB. They are widely used and perform well in the stochastic cascading bandits setting. Considering that our setting involves adversarial corruptions, we also compare our algorithms with the adversarial cascading bandits algorithm RBA. To evaluate the performance of the algorithms, we use the cumulative regret defined in Eq.(\ref{nips:regret}) as the evaluation metric, and all reported results are averaged over ten independent trials. We introduce the detailed method of generating datasets in the Appendix. 

\vspace{-0.5cm}
\subsection{Experiments in Synthetic Datasets}
\label{sec:synthetic}
For the synthetic datasets, we randomly generate $L = 500$ items, whose attraction probabilities are uniformly drawn in the interval $(0, 0.5)$. At each round $t$, the agent recommends $K=5$ items to the user. We set the total horizon $T= 1,000,000$. To introduce adversarial corruptions, we follow a similar approach as in ~\cite{bogunovic2021stochastic}.
The adversary corrupts the feedback by selecting the item with the lowest attraction probability as the target item. If the clicked item is not the target item, the adversary can modify the feedback of the clicked item to $0$. By adopting a ``Periodical" corruption mechanism~\cite{li2019cascading}, the adversary repeats the following process: (1) corrupt $t_{1}$ rounds, (2) keep the following $t_{2}$ rounds intact. We set $t_{1} = 10,000$ and $t_{2} = 90,000$ in this particular experiment. 
The results on the synthetic dataset are shown in Figure \ref{fig:synthetic}. We can find that our algorithms outperform all baselines, indicating the effectiveness of our approaches. In addition, CascadeRKC performs better than CascadeRAC. This can be attributed to the fact that CascadeRKC can leverage the information about the known corruption level, allowing it to make more informed decisions, which also matches our theoretical findings. Notice that the periodical corruptions do not significantly disrupt the performance of our algorithms, since both CascadeRKC and CascadeRAC are elimination-based algorithms; once an item is eliminated, it will not be selected in the future. We also find that the RBA algorithm, which is designed for the adversarial setting, does not perform well in the experiments. The reason is that, in our scenario, the feedback in most rounds is still stochastic and the RBA algorithm incurs a $\sqrt{T}$-level regret in stochastic environments.
\subsection{Experiments in the Real-world Datasets}
In this subsection, we choose three real-world datasets: Movielens~\cite{harper2015movielens}, Yelp\footnote{https://www.yelp.com/dataset}, and Yandex\footnote{https://www.kaggle.com/c/yandex-personalized-web-search-challenge}, which are commonly used in recommendation system experiments in Section \ref{sec:synthetic}. We will describe how to generate the data for the cascade model from the real-world datasets in the Appendix. We apply the same corruption method as the synthetic experiments. The results on real-world datasets are shown in Figure \ref{fig: real regret}. Our algorithms outperform other methods and CascadeRKC still performs better than CascadeRAC. Notice that the Yelp dataset, which is more sparse than other datasets, poses a greater challenge for the algorithms to converge. As a result, the regret scale in the Yelp dataset is larger than that in the Movielens and Yandex datasets. In the Movielens dataset, where the problems are relatively easier, some baselines such as CascadeKL-UCB and RBA also exhibit reasonable performance. However, they still fall short compared to our algorithms. 

\subsection{Experiments with Different Attraction Gaps}
To study how the attraction probability gaps between optimal and sub-optimal items affect the performance of the proposed methods, we conduct the following synthetic experiments. Let all the $K$ optimal items have the attraction probability $w = w_{1}$, and all sub-optimal items have the attraction probability $w = w_{2}$. Here we consider three cases, Case 1: $w_{1} = 0.2$ and $w_{2} = 0.1$; Case 2: $w_{1} = 0.3$ and $w_{2} = 0.1$; Case 3: $w_1=0.5$ and $w_2 = 0.1$. The corresponding values of $\Delta$ are $0.1$, $0.2$, $0.4$, respectively.

The results are provided in Figure \ref{fig: synthetic delta}. Our algorithms outperform all the baselines in all settings. The figures show that along with the increase of $\Delta$, the cumulative regret of algorithms decreases. This matches our theoretical result that $\Delta$ is in the denominator in the regret upper bound. 

\begin{figure}[t]
    \centering
    {
    \includegraphics[scale=0.33]{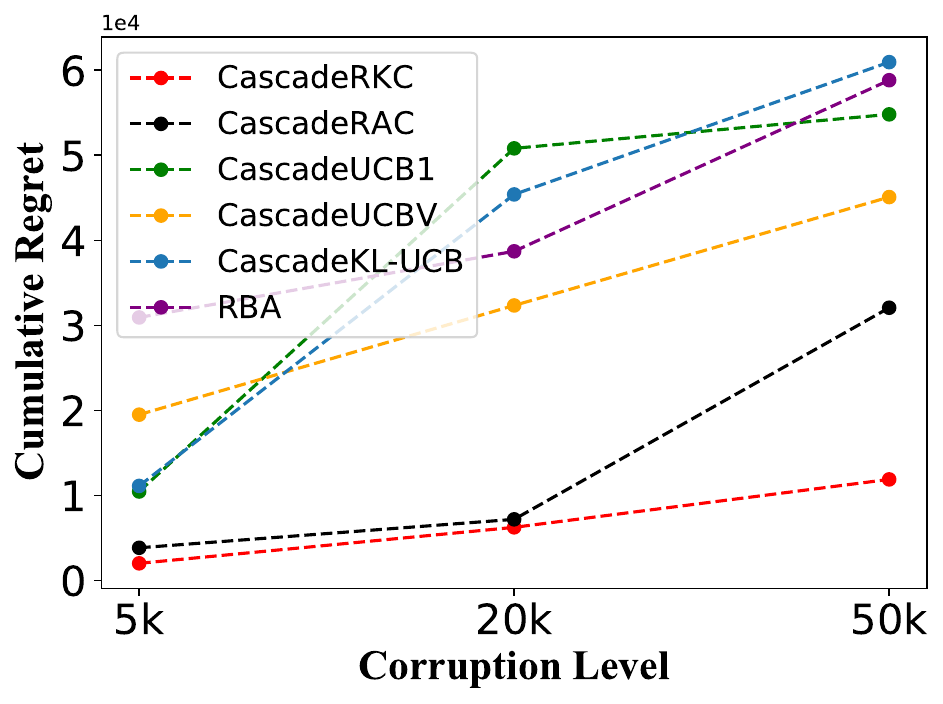}}
    \caption{Cumulative regret under different corruption levels in the synthetic dataset.}
    \label{fig:synthetic_corruption}
\end{figure}
\begin{figure}[t]
    \centering
    {
    \includegraphics[scale=0.33]{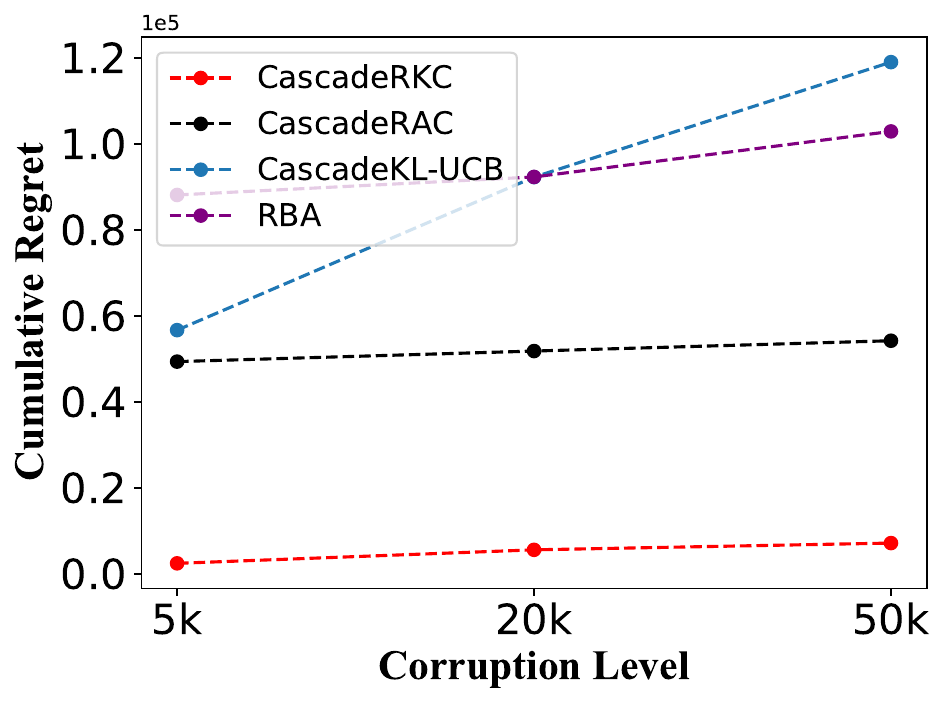}}
    \caption{Cumulative regret under different corruption levels in the Movielens dataset.} 
    \label{fig:movielens_corruption}
\end{figure}
\subsection{Experiments with Different Corruption Levels}
To explore the tolerance limit of our algorithms to the adversarial corruptions, we conduct the experiments in the synthetic dataset and Movielens dataset with different corruption levels. We adopt the ``Periodic" corruption mechanism described in section 6.1 and consider three cases, Case 1: $t_1=5,000$ and $t_2 = 95000$; Case 2: $t_1=20,000$ and $t_2 = 80,000$; Case 3: $t_1=50,000$ and $t_2 = 50,000$. In the synthetic dataset, we present the performance of all baselines to show the advantage of our algorithms, while in the Movielens dataset, we do not present CascadeUCB1 and CascadeUCBV as their performance is worse than that of other methods too much.

The results in the synthetic dataset are shown in Figure \ref{fig:synthetic_corruption} and the results in the Movielens dataset are provided in Figure \ref{fig:movielens_corruption}. We choose $T=1,000,000$ for the synthetic dataset and $T=2,000,000$ for the the Movielens dataset. We can see that in all settings the proposed CascadeRAC and CascadeRKC algorithms significantly outperform the baselines with significant advantages. In addtion, when the corruption level grows, the cumulative regret of our algorithms increases more slowly than that of baselines. These results also verify the robustness and good performance of our algorithms in various settings.

\subsection{Experiments with Different Corruption Mechanisms}
\begin{table}[t] \small
\label{tab:mechanism}
\centering
    \begin{tabular}{|c|c|c|c|c|}
    \hline
    & \multicolumn{2}{|c|}{Synthetic} & \multicolumn{2}{|c|}{Movielens}\\
    \hline
     & Periodic & Early & Periodic & Early\\
    \hline
    CascadeRKC &  2,499 & 68,876& 4,246&24,327\\
    \hline
    CascadeRAC &  4,930 & 72,468 & 51,246&66,342\\ 
    \hline
    CascadeUCB & 23,290 & 134,457 & 208,507&114,811 \\
    \hline
    CascadeUCBV & 27,311 & 89,612 &403,174 & 314,875\\
    \hline
    CascadeKL-UCB & 29,371 & 97,822&79,645& 18,6502 \\
    \hline
    RBA & 31,318 & 81,415& 89,039& 92,246\\
    \hline
    \end{tabular}
    \vspace{0.2cm}
    \caption{Cumulative regret with different corruption mechanisms in the synthetic and Movielens datasets.}
    \label{tab:mechanism}
\end{table}
In previous experiments, we assume the adversary attacks periodically. In this section, we conduct experiments in the synthetic and Movielens dataset with another corruption mechanism. Following \cite{bogunovic2021stochastic}, we assume the adversary attacks the first $100,000$ rounds in the synthetic dataset and $200,000$ rounds in the Movielens dataset, and leaves the remaining rounds intact. We call this mechanism an ``Early" mechanism as it puts all the corruptions to the early phase of the total horizon. Notice that though the total corrupted rounds are equal to the corresponding previous experiments, this mechanism may have a larger influence on the agent during the exploration phase.

We list the results in Table \ref{tab:mechanism}. We can see that putting all the corruptions at the early phase of the whole learning period can have a large influence on most algorithms in both the synthetic and Movielens dataset. However, CascadeRKC and CascadeRAC still perform better than the baselines and their advantages are even larger. These results exhibit the robustness of our algorithms under different corruption mechanisms.  

\section{Conclusion}
In this work, we first formulate the novel and challenging CBAC problem, where an adaptive adversary can manipulate the user feedback of a learning agent in cascading bandits to make it recommend sub-optimal items. To tackle this problem, we first design a novel position-based elimination algorithm (PBE) for the cascading bandits, which improves over the conventional active arm elimination method. Combining PBE and the idea of maintaining multiple instances to defend corruptions, we propose two robust algorithms, called \textit{CascadeRKC} and \textit{CascadeRAC}, which can resist different levels and mechanisms of corruption when the corruption level is known and agnostic. We provide sound theoretical guarantees for our algorithms, showing they can achieve $O(\log(T))$ regret bounds. We also conduct extensive experiments on both synthetic and real-world datasets to demonstrate the effectiveness and robustness of our algorithms under various settings. In the future, we will study how to design robust algorithms for online learning to rank with other click models such as position-based models.

\newpage
\bibliographystyle{unsrtnat}
\bibliography{arxiv_reference}  

\begin{thebibliography}{38}
\providecommand{\natexlab}[1]{#1}
\providecommand{\url}[1]{\texttt{#1}}
\expandafter\ifx\csname urlstyle\endcsname\relax
  \providecommand{\doi}[1]{doi: #1}\else
  \providecommand{\doi}{doi: \begingroup \urlstyle{rm}\Url}\fi

\bibitem[Kveton et~al.(2015{\natexlab{a}})Kveton, Szepesvari, Wen, and Ashkan]{kveton2015cascading}
Branislav Kveton, Csaba Szepesvari, Zheng Wen, and Azin Ashkan.
\newblock Cascading bandits: Learning to rank in the cascade model.
\newblock In \emph{International conference on machine learning}, pages 767--776. PMLR, 2015{\natexlab{a}}.

\bibitem[Combes et~al.(2015)Combes, Magureanu, Proutiere, and Laroche]{combes2015learning}
Richard Combes, Stefan Magureanu, Alexandre Proutiere, and Cyrille Laroche.
\newblock Learning to rank: Regret lower bounds and efficient algorithms.
\newblock In \emph{Proceedings of the 2015 ACM SIGMETRICS international conference on measurement and modeling of computer systems}, pages 231--244, 2015.

\bibitem[Cao et~al.(2007)Cao, Qin, Liu, Tsai, and Li]{cao2007learning}
Zhe Cao, Tao Qin, Tie-Yan Liu, Ming-Feng Tsai, and Hang Li.
\newblock Learning to rank: from pairwise approach to listwise approach.
\newblock In \emph{Proceedings of the 24th international conference on Machine learning}, pages 129--136, 2007.

\bibitem[Trotman(2005)]{trotman2005learning}
Andrew Trotman.
\newblock Learning to rank.
\newblock \emph{Information Retrieval}, 8:\penalty0 359--381, 2005.

\bibitem[Lattimore et~al.(2018)Lattimore, Kveton, Li, and Szepesvari]{lattimore2018toprank}
Tor Lattimore, Branislav Kveton, Shuai Li, and Csaba Szepesvari.
\newblock Toprank: A practical algorithm for online stochastic ranking.
\newblock \emph{Advances in Neural Information Processing Systems}, 31, 2018.

\bibitem[Zhong et~al.(2021)Zhong, Chueng, and Tan]{zhong2021thompson}
Zixin Zhong, Wang~Chi Chueng, and Vincent~YF Tan.
\newblock Thompson sampling algorithms for cascading bandits.
\newblock \emph{The Journal of Machine Learning Research}, 22\penalty0 (1):\penalty0 9915--9980, 2021.

\bibitem[Craswell et~al.(2008)Craswell, Zoeter, Taylor, and Ramsey]{craswell2008experimental}
Nick Craswell, Onno Zoeter, Michael Taylor, and Bill Ramsey.
\newblock An experimental comparison of click position-bias models.
\newblock In \emph{Proceedings of the 2008 international conference on web search and data mining}, pages 87--94, 2008.

\bibitem[Kveton et~al.(2015{\natexlab{b}})Kveton, Wen, Ashkan, and Szepesvari]{kveton2015combinatorial}
Branislav Kveton, Zheng Wen, Azin Ashkan, and Csaba Szepesvari.
\newblock Combinatorial cascading bandits.
\newblock \emph{Advances in Neural Information Processing Systems}, 28, 2015{\natexlab{b}}.

\bibitem[Li and De~Rijke(2019)]{li2019cascading}
Chang Li and Maarten De~Rijke.
\newblock Cascading non-stationary bandits: Online learning to rank in the non-stationary cascade model.
\newblock \emph{arXiv preprint arXiv:1905.12370}, 2019.

\bibitem[Lykouris et~al.(2018)Lykouris, Mirrokni, and Paes~Leme]{lykouris2018stochastic}
Thodoris Lykouris, Vahab Mirrokni, and Renato Paes~Leme.
\newblock Stochastic bandits robust to adversarial corruptions.
\newblock In \emph{Proceedings of the 50th Annual ACM SIGACT Symposium on Theory of Computing}, pages 114--122, 2018.

\bibitem[Gupta et~al.(2019)Gupta, Koren, and Talwar]{gupta2019better}
Anupam Gupta, Tomer Koren, and Kunal Talwar.
\newblock Better algorithms for stochastic bandits with adversarial corruptions.
\newblock In \emph{Conference on Learning Theory}, pages 1562--1578. PMLR, 2019.

\bibitem[Radlinski et~al.(2008)Radlinski, Kleinberg, and Joachims]{radlinski2008learning}
Filip Radlinski, Robert Kleinberg, and Thorsten Joachims.
\newblock Learning diverse rankings with multi-armed bandits.
\newblock In \emph{Proceedings of the 25th international conference on Machine learning}, pages 784--791, 2008.

\bibitem[Liu et~al.(2021)Liu, Li, and Li]{liu2021cooperative}
Junyan Liu, Shuai Li, and Dapeng Li.
\newblock Cooperative stochastic multi-agent multi-armed bandits robust to adversarial corruptions.
\newblock \emph{arXiv preprint arXiv:2106.04207}, 2021.

\bibitem[Lu et~al.(2021)Lu, Wang, and Zhang]{lu2021stochastic}
Shiyin Lu, Guanghui Wang, and Lijun Zhang.
\newblock Stochastic graphical bandits with adversarial corruptions.
\newblock In \emph{Proceedings of the AAAI Conference on Artificial Intelligence}, volume~35, pages 8749--8757, 2021.

\bibitem[Ding et~al.(2022)Ding, Hsieh, and Sharpnack]{ding2022robust}
Qin Ding, Cho-Jui Hsieh, and James Sharpnack.
\newblock Robust stochastic linear contextual bandits under adversarial attacks.
\newblock In \emph{International Conference on Artificial Intelligence and Statistics}, pages 7111--7123. PMLR, 2022.

\bibitem[Wang et~al.(2024)Wang, Xie, Yu, Li, and Lui]{wang2024online}
Zhiyong Wang, Jize Xie, Tong Yu, Shuai Li, and John Lui.
\newblock Online corrupted user detection and regret minimization.
\newblock \emph{Advances in Neural Information Processing Systems}, 36, 2024.

\bibitem[Even-Dar et~al.(2006)Even-Dar, Mannor, Mansour, and Mahadevan]{even2006action}
Eyal Even-Dar, Shie Mannor, Yishay Mansour, and Sridhar Mahadevan.
\newblock Action elimination and stopping conditions for the multi-armed bandit and reinforcement learning problems.
\newblock \emph{Journal of machine learning research}, 7\penalty0 (6), 2006.

\bibitem[Zoghi et~al.(2017)Zoghi, Tunys, Ghavamzadeh, Kveton, Szepesvari, and Wen]{zoghi2017online}
Masrour Zoghi, Tomas Tunys, Mohammad Ghavamzadeh, Branislav Kveton, Csaba Szepesvari, and Zheng Wen.
\newblock Online learning to rank in stochastic click models.
\newblock In \emph{International conference on machine learning}, pages 4199--4208. PMLR, 2017.

\bibitem[Vial et~al.(2022)Vial, Sanghavi, Shakkottai, and Srikant]{vial2022minimax}
Daniel Vial, Sujay Sanghavi, Sanjay Shakkottai, and R~Srikant.
\newblock Minimax regret for cascading bandits.
\newblock \emph{arXiv preprint arXiv:2203.12577}, 2022.

\bibitem[Li et~al.(2016)Li, Wang, Zhang, and Chen]{li2016contextual}
Shuai Li, Baoxiang Wang, Shengyu Zhang, and Wei Chen.
\newblock Contextual combinatorial cascading bandits.
\newblock In \emph{International conference on machine learning}, pages 1245--1253. PMLR, 2016.

\bibitem[Li and Zhang(2018)]{li2018online}
Shuai Li and Shengyu Zhang.
\newblock Online clustering of contextual cascading bandits.
\newblock In \emph{Proceedings of the AAAI Conference on Artificial Intelligence}, volume~32, 2018.

\bibitem[Zong et~al.(2016)Zong, Ni, Sung, Ke, Wen, and Kveton]{zong2016cascading}
Shi Zong, Hao Ni, Kenny Sung, Nan~Rosemary Ke, Zheng Wen, and Branislav Kveton.
\newblock Cascading bandits for large-scale recommendation problems.
\newblock \emph{arXiv preprint arXiv:1603.05359}, 2016.

\bibitem[Li et~al.(2019{\natexlab{a}})Li, Lattimore, and Szepesv{\'a}ri]{li2019online}
Shuai Li, Tor Lattimore, and Csaba Szepesv{\'a}ri.
\newblock Online learning to rank with features.
\newblock In \emph{International Conference on Machine Learning}, pages 3856--3865. PMLR, 2019{\natexlab{a}}.

\bibitem[Li et~al.(2020)Li, Feng, and Rijke]{li2020cascading}
Chang Li, Haoyun Feng, and Maarten~de Rijke.
\newblock Cascading hybrid bandits: Online learning to rank for relevance and diversity.
\newblock In \emph{Proceedings of the 14th ACM Conference on Recommender Systems}, pages 33--42, 2020.

\bibitem[Kapoor et~al.(2019)Kapoor, Patel, and Kar]{kapoor2019corruption}
Sayash Kapoor, Kumar~Kshitij Patel, and Purushottam Kar.
\newblock Corruption-tolerant bandit learning.
\newblock \emph{Machine Learning}, 108\penalty0 (4):\penalty0 687--715, 2019.

\bibitem[Zimmert and Seldin(2021)]{zimmert2021tsallis}
Julian Zimmert and Yevgeny Seldin.
\newblock Tsallis-inf: An optimal algorithm for stochastic and adversarial bandits.
\newblock \emph{The Journal of Machine Learning Research}, 22\penalty0 (1):\penalty0 1310--1358, 2021.

\bibitem[Agarwal et~al.(2021)Agarwal, Agarwal, and Patil]{agarwal2021stochastic}
Arpit Agarwal, Shivani Agarwal, and Prathamesh Patil.
\newblock Stochastic dueling bandits with adversarial corruption.
\newblock In \emph{Algorithmic Learning Theory}, pages 217--248. PMLR, 2021.

\bibitem[Hajiesmaili et~al.(2020)Hajiesmaili, Talebi, Lui, Wong, et~al.]{hajiesmaili2020adversarial}
Mohammad Hajiesmaili, Mohammad~Sadegh Talebi, John Lui, Wing~Shing Wong, et~al.
\newblock Adversarial bandits with corruptions: Regret lower bound and no-regret algorithm.
\newblock \emph{Advances in Neural Information Processing Systems}, 33:\penalty0 19943--19952, 2020.

\bibitem[Li et~al.(2019{\natexlab{b}})Li, Lou, and Shan]{li2019stochastic}
Yingkai Li, Edmund~Y Lou, and Liren Shan.
\newblock Stochastic linear optimization with adversarial corruption.
\newblock \emph{arXiv preprint arXiv:1909.02109}, 2019{\natexlab{b}}.

\bibitem[He et~al.(2022)He, Zhou, Zhang, and Gu]{he2022nearly}
Jiafan He, Dongruo Zhou, Tong Zhang, and Quanquan Gu.
\newblock Nearly optimal algorithms for linear contextual bandits with adversarial corruptions.
\newblock \emph{arXiv preprint arXiv:2205.06811}, 2022.

\bibitem[Dai et~al.(2024)Dai, Wang, Xie, Yu, and Lui]{dai2024online}
Xiangxiang Dai, Zhiyong Wang, Jize Xie, Tong Yu, and John~CS Lui.
\newblock Online learning and detecting corrupted users for conversational recommendation systems.
\newblock \emph{IEEE Transactions on Knowledge and Data Engineering}, 2024.

\bibitem[Bogunovic et~al.(2021)Bogunovic, Losalka, Krause, and Scarlett]{bogunovic2021stochastic}
Ilija Bogunovic, Arpan Losalka, Andreas Krause, and Jonathan Scarlett.
\newblock Stochastic linear bandits robust to adversarial attacks.
\newblock In \emph{International Conference on Artificial Intelligence and Statistics}, pages 991--999. PMLR, 2021.

\bibitem[Harper and Konstan(2015)]{harper2015movielens}
F~Maxwell Harper and Joseph~A Konstan.
\newblock The movielens datasets: History and context.
\newblock \emph{Acm transactions on interactive intelligent systems (tiis)}, 5\penalty0 (4):\penalty0 1--19, 2015.

\bibitem[Shi and Shen(2021)]{shi2021federated}
Chengshuai Shi and Cong Shen.
\newblock Federated multi-armed bandits.
\newblock In \emph{Proceedings of the AAAI Conference on Artificial Intelligence}, volume~35, pages 9603--9611, 2021.

\bibitem[Xie et~al.(2021)Xie, Yu, Zhao, and Li]{xie2021comparison}
Zhihui Xie, Tong Yu, Canzhe Zhao, and Shuai Li.
\newblock Comparison-based conversational recommender system with relative bandit feedback.
\newblock In \emph{Proceedings of the 44th International ACM SIGIR Conference on Research and Development in Information Retrieval}, pages 1400--1409, 2021.

\bibitem[Katariya et~al.(2017)Katariya, Kveton, Szepesv{\'a}ri, Vernade, and Wen]{katariya2017bernoulli}
Sumeet Katariya, Branislav Kveton, Csaba Szepesv{\'a}ri, Claire Vernade, and Zheng Wen.
\newblock Bernoulli rank-$1 $ bandits for click feedback.
\newblock \emph{arXiv preprint arXiv:1703.06513}, 2017.

\bibitem[Wu et~al.(2021)Wu, Zhao, Yu, Li, and Li]{wu2021clustering}
Junda Wu, Canzhe Zhao, Tong Yu, Jingyang Li, and Shuai Li.
\newblock Clustering of conversational bandits for user preference learning and elicitation.
\newblock In \emph{Proceedings of the 30th ACM International Conference on Information \& Knowledge Management}, pages 2129--2139, 2021.

\bibitem[Chuklin et~al.(2015)Chuklin, Markov, and Rijke]{chuklin2015click}
Aleksandr Chuklin, Ilya Markov, and Maarten~de Rijke.
\newblock Click models for web search.
\newblock \emph{Synthesis lectures on information concepts, retrieval, and services}, 7\penalty0 (3):\penalty0 1--115, 2015.

\end{thebibliography}






\section{Appendix}
\section{Proof of Lemma.\ref{lemma:slow corruption}}

\begin{proof}
 Given the fact that the expectation of the corruption in the slower instance $S$ will be a constant, we need to bound the variance of the corruption to get the possible highest corruption. We use $G_{t}(a)$ to denote the corruption at $t$ for an item $a$, and $H_{t}$ to denote the historical information till round $t$. Conditioned on selecting the instance $S$, let $a_{S,t}$ be one of the items that will be selected by the algorithm.  Then we define \begin{equation*}
     X_{t} = G_{t}(a_{S,t}) - \EE\left[G_{t}(a_{S,t}) | H_{t-1}\right],
 \end{equation*} which is a martingale sequence, representing the deviation of the actual corruption incurred by item $a_{S,t}$ from its conditional expectation. And we have
 \begin{align*}
     \mathbb{E}\left[G_{t}(a_{S,t}) | H_{t-1}\right] = \frac{1}{C}C_{a_{S,t}} + 0 = \frac{C_{a_{S,t}}}{C},
 \end{align*}
 where $C_{a_{S,t}}$ denote the corruption that the adversary selects for $a_{S,t}$, and the equalities hold as $G_{a_{S,t}}$ equals to $C_{a_{S,t}}$ with probability $1/C$, and otherwise $G_{a_{S,t}}=0$. Hence, we can derive that, 
\begin{align*}
    \mathbb{E}[X_{t}^2|X_1,\cdots,X_{t-1}] &= \frac{1}{C}\big(C_{a_{S,t}}-\frac{C_{a_{S,t}}}{C}\big)^2 + \big(1-\frac{1}{C}\big)\big(\frac{C_{a_{S,t}}}{C}\big)^2\\ 
    &\le \frac{2C_{a_{S,t}}}{C}, 
\end{align*}
which holds by the fact that $C_{a_{S,t}}$ can only be $1$ or $0$ in the cascade model and $C_{a_{S,t}}\le C$. 
 Applying LemmaB.1 in \cite{lykouris2018stochastic}, we have, with probability at least $1-\delta_1$,\begin{align*}
     \sum_{t\in[T]}X_t &\le \log(1/\delta_1) + (e-2)\cdot\sum_{t\in[T]}  \mathbb{E}[X_{t}^2|X_1,\cdots,X_{t-1}] \\
    & \le \log(1/\delta_1) + (e-2)\cdot2\sum_{t\in[T]}\frac{C_{a_{S,t}}}{C}\\
     &\le \log(1/\delta_1) + 2. 
 \end{align*}
 The third inequality holds by $\sum_{t}C_{a_{S,t}}\le C$ by the definition of C. Thus, we have \begin{align*}
     \sum G_{t}(a) &= \sum X_{t} + \EE\left[{G_{t}(a) | H_t}\right] \\
     &\le \log(1/\delta_1) + 2 + \frac{C_{a_{S,t}}}{C}\\
     &\le \log(1/\delta_1) + 3
 \end{align*}, where the first inequality holds as $C_{a_{S,t}}\le C$.
 
\end{proof}
\section{Proof of Lemma.\ref{lemma:slow times}}
\begin{proof}
     Let $\hat{w}_{o}^{S}(a)$ represent the estimation of $w(a)$ from the stochastic feedbacks, then by the Hoeffding inequality, at some timestep $t$, there exits $\delta_{2}\in[0,1]$ such that, with probability at least $1-\delta^{'}$:
\begin{equation}
    \left|\hat{w}_{o}^{S}(a) - w(a)\right| \le \sqrt{\frac{\log\left(2/\delta_{2}\right)}{T^{S}(a)}}.
\end{equation}
Selecting $\delta_{2} = \delta^{'}/LT$, by a union bound argument, we have, for any $a \in E$ and $t\in[T]$,
\begin{align*}
   \left|\hat{w}_{o}^{S}(a) - w(a)\right| \le \sqrt{\frac{\log\left(2/\delta^{'}\right)}{T^{S}(a)}},
\end{align*}
with probability at least $1-\delta_{2}$. Using $C_{s}$ to denote the corruptions in the $S$ instance, the corruption will make at most a $C_{s}/T^{s}(a)$ disturbance to the estimation of $a$. Then \begin{align*}
    \hat{w}^{S}(a_{k}) &\ge \hat{w}_{ o}^{S}(a_{k}) - \frac{C_{s}}{T^{S}(a_{k})} \\&\ge w(a_{k}) - \sqrt{\frac{\log\left(2LT/\delta_2\right)}{T^{S}(a_{k})}} - \frac{C_{s}}{T^{s}(a_{k})}.
\end{align*}And, similarly, \begin{align*}
    \hat{w}^{S}(e) &\le \hat{w}^{S}_{o}(e) + \frac{C_{s}}{T^{s}(e)} \\&\le w(e) + \sqrt{\frac{\log\left(2LT/\delta_2\right)}{T^{s}(e)}} + \frac{C_{s}}{T^{s}(e)}.
\end{align*} As suggested in Lemma.\ref{lemma:slow corruption}, $C_{s} \le \log\left(1/\delta_1\right) + 3$, which is less than  $2\log\left(8LT/\delta_2\right)$ in the numerator of the second term in our $wd^{S}(\cdot)$ by taking $\delta_2=\delta_1$. Thus, we have
\begin{align*}
    \hat{w}^{S}(a_{k}) \ge w(a_k) - wd^{S}(a_k), \hat{w}^{S}(e)\le w(e) + wd^{S}(e).
\end{align*} 
And \begin{align*}
    \hat{w}^{S}(e) - \hat{w}^{S}(a_k) &\le wd^{S}(a_k) + wd^{S}(e) - \Delta_{e,k}\\
    &\le wd^{S}(a_k) + wd^{S}(e)
\end{align*}
implies that the optimal items will never be eliminated. Moreover, as in the slower instance $S$, a sub-optimal arm $e$ will be eliminated from position $k$ when $\hat{w}^{S}(a_{k}) - wd^{S}(a_{k}) > \hat{w}^{S}(e) + wd^{S}(e)$ according to the elimination rule, then with the $T^{S}(a)$ in Eq.(\ref{eq:slow times}) played times for $e$ (also for $a_k$ as $a_k$ is not eliminated), $\hat{w}_{k}^{S}(a_{k}) - wd^{S}(a_{k}) > \hat{w}_{k}^{S}(e) + wd^{S}(e)$ will be satisfied, and the item $e$ will be eliminated for position $k$. 
\end{proof}

\section{Proof of Theorem.3}
\begin{proof}
To derive a high-probability upper bound of the regret, we first derive the expected regret and then analyze its variance. The regret of CascadeRKC comes from the $F$ and $S$ instances. We first bound the regret generated in the $S$ instance. By Theorem 1 in \citep{kveton2015cascading}, the expectation of immediate regret $R_t$ at period $t$ conditioned on $\mathcal{H}_t$ can be upper bounded by
\begin{equation*}
    \EE[R_t|\mathcal{H}_t] \le \sum_{e=K+1}^{L}\sum_{k=1}^{K}\Delta_{e,k} \EE[\mathbb{I}\{\mathcal{E}(e,k,t)\}],
\end{equation*}
where $\mathbb{I}$ denotes the indicator function and $\mathcal{E}(e,k,t)$ represents the event that 
item $e$ is chosen instead of item $k$ at time
$t$, and that item $e$ is observed. Then the total expected regret can be bounded by
\begin{align*}
    \EE[R(T)] &\le \sum_{t=1}^{T}\EE[ \EE[R_t|\mathcal{H}_t] ] \\
    &\le \EE\big[\sum_{t=1}^{T}\sum_{e=K+1}^{L}\sum_{k=1}^{K}\Delta_{e,k} \EE[\mathbb{I}\{\mathcal{E}(e,k,t)\}]\big].
\end{align*}
Notably, for each pair $(e,k)$, we have \begin{align*}
\EE\big[\sum_{t=1}^{T}\Delta_{e,k}\EE[\mathbb{I}\{\mathcal{E}(e,k,t)\}]\big] &\le (1-\delta_2)T^{S}(e)\Delta_{e,k} + \delta_2 \cdot T\Delta_{e,k}\\
    &\le (1-\delta_2)\frac{18\log(8LT/\delta_2)}{\Delta_{e,k}} + \delta_2T\Delta_{e,k} \\
    &\lesssim \frac{18\log(8LT/\delta_2)}{\Delta_{e,k}}.
\end{align*}
The first two inequalities hold by  Lemma.\ref{lemma:slow times}, the last inequality follows by taking 
$\delta_2=1/T$. Now we need to calculate the variance to get the possible upper bound. By the Hoeffding inequality, the gap between the empirical cumulative reward of arm $e$ and its expectation is at most $\sqrt{T^{S}(e)\log(2LT/\delta^{''})}$ with probability $1-\delta^{''}$. With:
\begin{equation*}
\begin{aligned}
    \sqrt{T^{S}(e)\log(2LT/\delta^{''})} &\le T^{S}(e)\sqrt{\frac{\log(2LT/\delta^{''})}{T^{S}(e)}}\\
    &\le T^{S}(e)\Delta_{e,k}\sqrt{\frac{\log(2LT/\delta^{''})}{18\log(8LT/\delta_2)}} \\
    & \le T^{S}(e)\Delta_{e,k} \\
    &\le \frac{18\log(8LT/\delta_2)}{\Delta_{e,k}},
\end{aligned}
\end{equation*}
where the last inequality holds by taking $\delta^{''}=\delta_2$.
With a similar argument to item $k$, we can get the regret caused by playing $e$ at $k$ position in $S$ can be upper bounded by $O(36\log(8LT/\delta_2))/\Delta_{e,k}$.

In the $F$ instance, in expectation the arm $e$ will appear at position $k$ by $L\times CT^{S}(e)$ rounds as every move in the slow active arm elimination occurs with
probability $1/C$ and, at least $1/L$ of these moves are plays of $e$ while it is still active. To obtain a high probability guarantee, observe that with probability at least $1-\delta_{4}$, we make one move at the slow arm elimination algorithm every $O\left(C\log(1/\delta_{4})\right)$ moves at the
fast arm elimination algorithm \citep{lykouris2018stochastic}. Taking the union bound $\delta^{'''} =LT^{S}(e)\times \delta_{4}$, we can get arm $e$ will be eliminated for position $k$ in $F$ at most when:
\begin{equation*}
\begin{aligned}
    T^{F}(e) &\le L \times C \times T^{S}(e) \times \log(1/\delta^{'''}) \\
    &\le \frac{18CL\left(\log(8LT/\delta)\right)^{2}}{\Delta_{e,k}^2},
\end{aligned}
\end{equation*}
where we use $\delta$ to unify all the uncertainties. 
Then following a similar analysis for the $S$ instance, and with the fact that there are $K$ positions and $L-K$ sub-optimal arms, we can complete the proof.
\end{proof}

\section{Proof of Theorem.4}
\begin{proof}
    We can divide the $\log(T)$ instances in \emph{CascadeRAC} into two parts and consider their regrets respectively: one part includes all the layers which satisfy: $2^{\ell} > C$; and the other part, in contrast, is consisted of the layers with $2^{\ell} \le C$. For the instances in the first part, their corruption levels are higher than $C$, and they are stochastic. With a similar analysis for the \textit{CascadeRKC}, the sub-optimal item $e$ will appear $O\big(\frac{\log(LT/\delta)}{\Delta_{e,k}^{2}}\big) $times at the $k$ position with a $\Delta_{e,k}$ regret at each appearance, then each layer in the first part can generate at most $O\big(\frac{\log(LT/\delta)}{\Delta_{e,k}}\big)$ regret for each sub-optimal item $e$ and position $k$. As there are most $\log(T)$ such layers, the regret for these instances can be bounded by $\sum_{e=K+1}^{L}K\log(T)\frac{\log(LT/\delta)}{\Delta_{e,K}}$, which is the second term in \emph{CascadeRAC}'s regret upper bound.

    As for the other part of the instances, we use a similar technique in the analysis of \emph{CascadeRKC} to bound their regrets. We first consider the minimum instance $\ell^{*}$ which satisfies $2^{\ell^{*}} > C$, and as the corruption level increases by a power of 2 in the instances, we can get $2^{\ell^{*}} \le 2C$. Then we can use a similar way in Lemma.\ref{lemma:slow corruption} and Lemma.\ref{lemma:slow times} to bound the played times of a sub-optimal item $e$ in the $\ell^{*}$ instance which is $O\left( \frac{\log(LT/\delta)}{\Delta_{e,k}^{2}} \right) $. The next step is to bound the played times of item $e$ in the instances $2^{\ell} \le C$ by taking a union bound argument like in the analysis of \textit{CascadeRKC}, which results in the first term in \emph{CascadeRAC}'s regret upper bound. Combining the result with the first part instances, we can complete the proof.  
\end{proof}

\section{The Generation of Real-world Datasets}
We use three real-world datasets: Movielens \citep{harper2015movielens}, Yelp\footnote{https://www.yelp.com/dataset}, and Yandex\footnote{https://www.kaggle.com/c/yandex-personalized-web-search-challenge}. The Movielens dataset consists of 2,113 users and 10,197 movies. The Yelp dataset comprises 4.7 million ratings from 1.18 million users for approximately 157,000 restaurants. For these two datasets, we select 1,000 users who rate most and 500 items with the most ratings \citep{shi2021federated,li2018online,xie2021comparison}. The Yandex click dataset is the largest public click collection, it contains 35 million search sessions, each of which may contain multiple search queries. Following \citep{katariya2017bernoulli}, we select $20$ frequent queries from the dataset. To maintain consistency with the other datasets, we select the top 500 items with the highest attraction probabilities within each query. The reported results on the Yandex dataset are averaged over these 20 queries.

In the Movielens and Yelp datasets, we first construct the feedback matrix $H^{1,000 \times 500}$ for the selected users and items according to the rating \citep{wu2021clustering}:
for each user-item pair, if the rating is greater than 3, the corresponding feedback is set to 1; otherwise, it is set to 0. At each round $t$, a user $i$ comes randomly, the agent recommends $K=5$ items. And if there is no corruption at this round, the user $i$ will receive the feedback $F^{i,j}$ by clicking item $j$. Similar to \citep{zong2016cascading}, our goal is to maximize the probability of recommending at least one attractive item in these two datasets. And since these two datasets are much large, we set the total horizon $T=2,000,000$ to allow enough time for convergence. In the Yandex dataset, following \citep{zoghi2017online}, we use PyClick~\citep{chuklin2015click} to learn the cascade model. We select $20$ queries, and each query contains $L = 500$ items, where each item is assigned a weight learned from the PyClick. We set $T=1,000,000$, and at each round, the agent recommends $K=5$ items to the user. In all the three real-world datasets, the corruption mechanism is the same as the synthetic dataset.

\end{document}